\documentclass[letterpaper, 10 pt, conference]{ieeeconf}  

\IEEEoverridecommandlockouts                              

\usepackage[utf8]{inputenc}
\usepackage[T1]{fontenc}
\usepackage{url}
\usepackage{booktabs}
\usepackage{float}
\usepackage{amsfonts}
\usepackage{nicefrac}
\usepackage{microtype}
\usepackage{amsmath}
\usepackage{makecell}
\usepackage{multirow}
\usepackage{amssymb}
\usepackage{pifont}
\usepackage[table]{xcolor}
\usepackage{graphicx}
\usepackage{algorithm}
\usepackage{algorithmic}
\usepackage[most]{tcolorbox}
\usepackage{hyperref}
\newcommand{\conf}[1]{\textcolor{gray}{\scriptsize [#1]}}

\definecolor{best}{RGB}{186,220,255}
\definecolor{second}{RGB}{255,218,185}

\title{\LARGE \bf
Goal2Pixel: Grounding Goals to Pixels for Vision-Language Navigation
}

\author{
    \textbf{Muyi Bao}$^{1,*,\dagger}$,
    \textbf{Yuxin Cai}$^{1,2,*}$,
    \textbf{Hang Xu}$^{1}$,
    \textbf{Zongtai Li}$^{1}$,
    \textbf{Jinxi He}$^{1}$, 
    \textbf{Jingfan Tang}$^{1}$\\
    \textbf{Chen Lv}$^{2}$,
    \textbf{Ji Zhang}$^{1}$,
    \textbf{Yaqi Xie}$^{1}$,
    \textbf{Wenshan Wang}$^{1}$ \\
    $^{1}$ Carnegie Mellon University 
    $^{2}$ Nanyang Technological University \quad \\
    $^{*}$ Equal contribution. 
    $^{\dagger}$ Corresponding author.
}

\begin{document}

\maketitle
\thispagestyle{empty}
\pagestyle{empty}

\begin{abstract}

Vision-language models (VLMs) have become a common foundation for vision-and-language navigation in continuous environments (VLN-CE). Yet most VLM-based methods cast navigation as low-level action prediction, an interface that is ambiguous, tied to short-horizon motion primitives, and inefficient due to repeated VLM querying. We propose \textbf{Goal2Pixel}, a pure pixel-based paradigm that reformulates VLN-CE as navigable pixel grounding. Rather than predicting actions, Goal2Pixel uses the image plane as a unified spatial interface between VLM reasoning and robot motion: the model predicts a visible navigable pixel, which is back-projected into a 3D waypoint for forward navigation. Non-forward actions are encoded by auxiliary left/right/bottom regions for turning and stopping.
To support long-horizon navigation, we propose ViKeyMem, a visibility-aware keyframe memory, achieving a favorable performance–efficiency trade-off while reducing training cost by over 50\%.
To adapt pretrained VLMs to navigable pixel grounding, we introduce semantic embeddings and coordinate-aware auxiliary losses.
Among methods trained without external data, Goal2Pixel achieves competitive state-of-the-art performance on R2R-CE and RxR-CE. On R2R-CE Val-Unseen, it achieves 54.1\% SR and 52.5\% SPL with only 7.75 VLM calls per episode. Compared with direct action prediction, Goal2Pixel improves SR by 21.2 percentage points while requiring 6$\times$ fewer VLM calls.
\href{https://anonymous.4open.science/w/Goal2Pixel-031D/}{Project Page}. 

\end{abstract}


\section{INTRODUCTION}
Vision-and-language navigation in continuous environments (VLN-CE) \cite{r2r-ce, rxr} requires an embodied agent to follow natural-language instructions and reach a target through fine-grained physical movements. Unlike static vision-language understanding, VLN-CE is a language-conditioned spatial decision-making problem: the agent must ground language in egocentric observations, track its progress, and continuously decide where to move next \cite{vln_bert, hamt, fgr2r, know_where}. Vision-language models (VLMs) \cite{internvl, qwen25vl} offer a natural foundation for this problem, and recent VLN-CE methods increasingly build on pretrained VLMs to improve instruction-following and generalization \cite{efficientVLN, Janusvln, Uni-navid, navid, nav-r1, Navila}. Yet the interface between this high-level reasoning and executable motion has barely evolved: most VLM-based methods still query the VLM at every timestep for low-level meta-actions such as \textit{turn left/right 15°, move forward 25 cm, or stop}.

This action-centric interface in VLN-CE  has three fundamental limitations. (i) Ambiguous action supervision: many distinct action sequences can reach the same goal, making the oracle trajectory an unnecessarily restrictive training target \cite{r2r-ce, krantz2021waypoint, narrowthegap}. (ii) Myopic decisions: optimizing for the next short-range movement instead of longer-horizon spatial reasoning \cite{krantz2021waypoint}. (iii) High inference cost: because each prediction advances the agent only a few centimeters or degrees, the VLM must be invoked dozens of times per episode \cite{navid, Janusvln, Streamvln, Uni-navid}. Therefore, the bottleneck is not only the capacity of the VLM but the interface through which its reasoning becomes motion.


We therefore ask: \textbf{can the image plane serve as a unified interface between VLM reasoning and executable motion in VLN-CE?}
Motivated by recent explorations of pixel-based navigation decisions~\cite{dualvln}, we propose \textbf{Goal2Pixel}, a \textbf{pure pixel-based navigation paradigm} that represents both forward and non-forward decisions within a unified image-space output. To unify the action space, we augment the image with auxiliary \emph{directive regions}, allowing turning and stopping decisions to be represented in the same pixel-coordinate space as forward navigation.
For forward navigation, the VLM predicts a visible navigable goal pixel in the current observation, which is back-projected into a 3D waypoint for low-level execution.
For non-forward decisions, the left, right, and bottom directive regions encode \texttt{Turn\_Left}, \texttt{Turn\_Right}, and \texttt{Stop}, respectively.
In this way, all high-level navigation decisions are expressed through a single pixel-coordinate output space.

To support long-horizon navigation, navigators require memory of instruction-relevant landmarks and visited regions, but existing VLM-based navigators feed 8--32 fixed-interval or uniformly sampled frames \cite{efficientVLN, dualvln, Janusvln, Uni-navid, navid, omninav, monodream, Navila}, introducing substantial visual redundancy and computational cost as trajectories grow longer. Therefore, we propose ViKeyMem, a visibility-aware keyframe memory that retains a frame only when the set of visible waypoints changes substantially and overlays the past trajectory on each retained frame, yielding only 4--5 keyframes per 100 timesteps. By filtering redundant history without additional memory modules, ViKeyMem provides a favorable performance--efficiency trade-off, outperforming conventional history methods on the long-horizon RxR benchmark while reducing training cost by over 50\%.

To make navigable pixel grounding effective with pretrained VLMs, we introduce two lightweight adaptations. 
\emph{First}, since the auxiliary directive regions and trajectory overlays are outside the VLM's pretraining distribution, we attach learnable semantic embeddings on these special tokens to help the model distinguish them from regular RGB tokens. 
\emph{Second}, to better align training with the geometry of pixel prediction, we add coordinate-aware auxiliary losses, including numeric and angular terms.

We evaluate Goal2Pixel on the continuous versions of Room-to-Room (R2R)~\cite{r2r-ce} and Room-across-Room (RxR)~\cite{rxr}. 
Compared with prior methods that also do not use external training data, Goal2Pixel achieves competitive state-of-the-art navigation performance.
Our ablation studies further show the pure pixel-based paradigm consistently outperforms action-based and hybrid action-pixel-based output designs. We further deploy Goal2Pixel zero-shot onboard a wheeled robot, demonstrating its feasibility for real-world navigation.

\textbf{Contributions:} (1) Goal2Pixel, a pure pixel-based interface that reformulates VLM-based VLN-CE from action prediction to image-space goal grounding; 
(2) ViKeyMem, a visibility-aware history representation for long-horizon navigation that reduces training cost by over 50\% while maintaining strong performance; 
(3) Lightweight adaptations, including semantic embeddings and coordinate-aware auxiliary losses;
(4) Goal2Pixel achieves 54.1/52.5 SR/SPL with the 2B model on R2R-CE and 48.1/44.7 with the 8B model on RxR-CE.. Compared with direct action prediction, our Goal2Pixel improves SR in R2R-CE by 21.2 percentage points and reduces the average number of VLM calls by about 6$\times$.

\section{RELATED WORK}
\label{sec:related_works}
\subsection{Vision-and-Language Navigation}

Early VLN benchmarks, such as Room-to-Room (R2R)~\cite{r2r-vln}, are built on discrete navigation graphs, where the agent moves between predefined viewpoints. Recent benchmarks extend VLN to continuous environments, including R2R-CE~\cite{r2r-ce} with fine-grained control and RxR-CE~\cite{rxr} with longer and more complex instructions. Existing VLN-CE methods can be broadly grouped into zero-shot approaches, non-VLM learning-based approaches, and recent VLM-based approaches. Zero-shot methods~\cite{boosting-zero-shot, instructnav, AO-Planner, CA-Nav, LaViRA} leverage foundation models for reasoning-based navigation, but often suffer from high inference cost and limited alignment with continuous control. 
Non-VLM learning-based methods~\cite{CMA, Navmorph, etpnav, g3d-lf, Cosmo, sim2real, sim2sim} train task-specific navigation policies, frequently simplifying continuous navigation through graphs, maps, waypoints, or low-level action prediction. Although effective, these formulations are limited by less generalization ability \cite{Navmorph, vint} and accumulate errors \cite{enp, minderror} during long-horizon execution.

\subsection{Output Paradigm}
Recent VLM-based navigation methods~\cite{navid, Mapnav, Navila, Janusvln, Streamvln} leverage pretrained vision-language backbones for embodied navigation, but most still follow an action-centric paradigm. Several works have explored alternative output interfaces. NavFoM~\cite{navfom} predicts relative 3D coordinates for cross-embodiment and cross-task unification, while OmniNav~\cite{omninav} predicts continuous-space waypoints for multi-task navigation. The most related work to ours is DualVLN~\cite{dualvln}, which also incorporates pixel prediction. 
DualVLN follows an Action-Pixel (Seq) paradigm, where the model first predicts an action-oriented decision, such as \texttt{turning}, \texttt{stopping}, or \texttt{Look\_Down}, and then pixel grounding is triggered only when \texttt{Look\_Down} is selected, requiring a second-stage prediction to determine the navigation target.
In contrast, Goal2Pixel uses directive regions to unify forward and non-forward decisions into a pure pixel paradigm, eliminating the sequential switch between action prediction and pixel grounding.
Our experiments show that the pure pixel paradigm consistently outperforms the Action-Pixel (Seq) paradigm in SR/SPL while requiring fewer VLM calls on both R2R-CE and RxR-CE.

\subsection{History Representation}
Historical observations are important for long-horizon VLN, as they preserve previously seen landmarks and decision points. Existing methods often construct visual history through simple frame selection strategies. Fixed-interval sampling \cite{efficientVLN, dualvln, Janusvln, Uni-navid, navid, omninav} keeps one frame every $N$ timesteps, but it can introduce redundant observations in visually similar regions while still providing an incomplete representation of the whole trajectory. Uniform sampling \cite{monodream, Navila} and multi-frequency sampling \cite{nav-r1,vln-r1} improve temporal coverage, yet they may still retain redundant frames in early stages and miss important intermediate observations between two sampled frames in later stages. 
Beyond simple frame sampling, several methods use more structured or compressed history representations \cite{Mapnav, navcot, navfom, Streamvln, intentNav}. While these designs improve history modeling, they often introduce additional modules or computational overhead. 
In contrast, ViKeyMem aims to preserve navigation-relevant context with a small number of keyframes without introducing additional modules or substantial computational overhead.

\section{Goal2Pixel}

\begin{figure*}[h]
    \centering
    \includegraphics[width=0.9\linewidth]{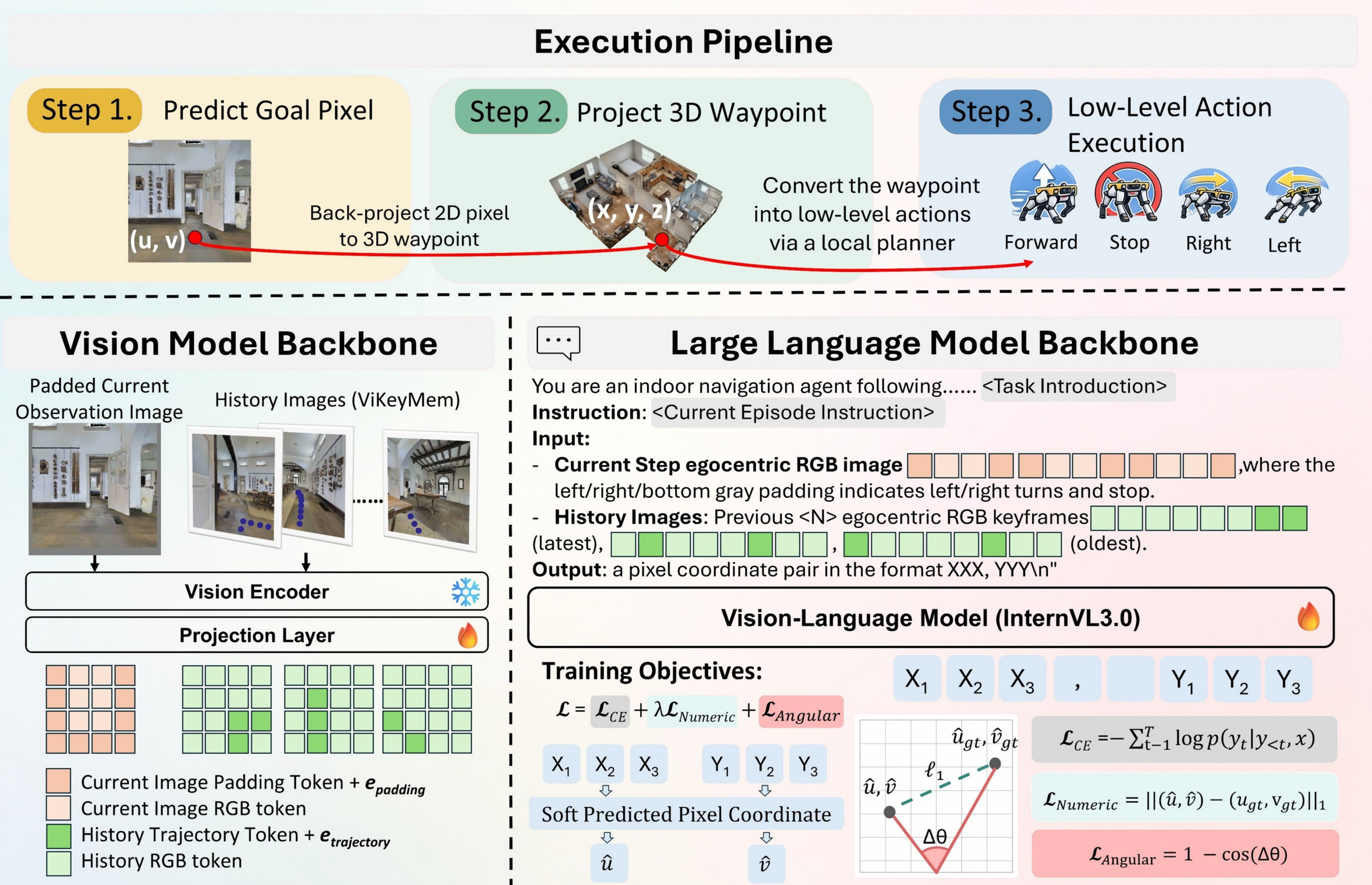}
    \caption{\textbf{Overview of Goal2Pixel.} 
The framework consists of a three-stage execution pipeline and a VLM-based prediction module that outputs goal pixels.
\textbf{Top: Execution pipeline.}
At each decision step, the VLM predicts a goal pixel $(u,v)$ (Step 1). Pixels in the RGB region are back-projected with depth and camera geometry to 3D waypoints $(x,y,z)$ (Step 2), which are converted into low-level actions by a local planner (Step 3). Pixels in auxiliary directive regions denote non-forward actions.
\textbf{Bottom: Vision-language model.} 
A VLM takes as input the instruction, the padded current egocentric RGB observation, and a compact history memory (Sec \ref{sec:vikeymem}), and outputs a navigable goal pixel. 
Semantic embeddings are introduced to distinguish regular RGB, auxiliary directive region token, and trajectory overlay tokens, and the model is trained with both token-level cross-entropy loss and auxiliary coordinate-aware losses.}
    \label{fig:pipeline}
\end{figure*}

\subsection{Vision-Language Navigation Task Definition}

Vision-and-Language Navigation in continuous environments (VLN-CE) requires an embodied agent to follow a natural language instruction $\mathcal{I}$. At each time step $t$, the agent receives an egocentric RGB observation $\mathcal{O}_t$, a synchronized depth observation $\mathcal{D}_t$, and its current pose $\mathbf{s}_t$, including the agent position and orientation. In the standard VLN-CE setting, the agent acts in a low-level action space $\mathcal{A}$ consisting of \texttt{Move\_Forward (25 cm)}, \texttt{Turn\_Left ($15^\circ$)}, \texttt{Turn\_Right ($15^\circ$)}, and \texttt{Stop}. After executing an action, the agent receives updated observations and pose information, and the process repeats until it outputs \texttt{Stop}. An episode succeeds if the agent stops within the instruction-specified target region.

\subsection{Data Collection}
We derive oracle trajectories from the training splits of R2R-CE~\cite{r2r-ce} and RxR-CE (English subset)~\cite{rxr} in MP3D~\cite{matterport3d}, totally constructing 2.64M training samples. Each sample is represented as a tuple $(\mathcal{I}, O_{\text{cur}}, O_{\text{hist}}, p)$, where $\mathcal{I}$ is the instruction, $O_{\text{cur}}$ is the padded current egocentric RGB image, $O_{\text{hist}}$ is the sequence of historical images constructed by ViKeyMem, and $p$ is the ground-truth pixel.

\subsubsection{Ground-Truth Pixel Construction.}
We define the target pixel $p$ by projecting the farthest visible future waypoint $w$ onto the current image. Visibility is verified by a project--reproject consistency check. 
Specifically, the waypoint $w$ is projected to a pixel $p$, and $p$ is then back-projected using its depth to obtain $\hat{w}$; $w$ is considered visible if $\lVert w-\hat{w}\rVert_2 < 0.6$ m. This encourages the model to predict a farther navigable goal.
To represent non-forward decisions within the same pixel-prediction space, we append 8-pixel-wide directive regions to the left and right of the RGB image and a 16-pixel-high region to the bottom, all filled with RGB value $(128,128,128)$.
We use these regions in two geometric fallback cases:
1) Task Completion: if the agent is within 1.0 m of the final destination, $p$ is assigned to the center pixel of the bottom auxiliary directive region to indicate \texttt{Stop}; 2) Out-of-View: if no future waypoint is visible, $p$ is assigned to the center pixel of the left or right auxiliary directive regions according to the mean egocentric direction of the next five waypoints, indicating \texttt{Turn\_Left} or \texttt{Turn\_Right}.

\subsubsection{History Construction with ViKeyMem} \label{sec:vikeymem} For each timestep $t$, we construct the history observations $O_{\text{hist}}$ from the observed trajectory prefix. For a historical frame at timestep $i<t$, subsequent waypoints refer only to the executed trajectory from $i$ to $t$, never to future oracle waypoints.

ViKeyMem selects keyframes according to the visibility of these subsequently observed waypoints from each candidate frame. A candidate frame is selected as a keyframe if it satisfies three conditions: (1) it is not visible from the most recently selected keyframe; (2) at least one subsequent waypoint is visible in the candidate's egocentric image; and (3) at least two distinct subsequent waypoints fall within the candidate's $45^\circ$ forward field of view. The first condition is the core visibility criterion, ensuring that a new keyframe is added only when the candidate provides information not covered by the previous keyframe. The latter two conditions ensure that the selected keyframe provides informative visual context for the subsequently executed trajectory. 

At inference time, the keyframe memory is updated retrospectively as new trajectory observations become available, using only information observed up to the current timestep. Because the selected keyframes are sparse, we overlay the agent's executed trajectory onto each keyframe by drawing blue dots along the trajectory, making the motion connection between distant observations explicit.

\subsection{Goal2Pixel Pipeline}\label{sec:Goal2Pixel}

The overall pipeline of Goal2Pixel is illustrated in Fig.~\ref{fig:pipeline}. 
Goal2Pixel casts navigation as a visually grounded goal prediction problem. 
Given the instruction, padded current observation, and navigation history, Goal2Pixel autoregressively outputs a pixel coordinate as text in the format ``XXX, YYY''.
This predicted pixel serves as a spatially grounded intermediate representation for navigation.
Specifically, if the predicted pixel lies in the regular RGB region, its corresponding depth is retrieved from $\mathcal{D}_t$ and back-projected through the camera geometry to obtain a 3D waypoint in the agent coordinate frame. A lightweight local planner, such as FMM planner \cite{LaViRA}, then tracks this waypoint and converts it into executable navigation actions.
If the predicted pixel falls into one of the predefined auxiliary directive regions, it is directly interpreted as \texttt{Turn\_Left}, \texttt{Turn\_Right}, or \texttt{Stop}. 
In this way, Goal2Pixel decouples high-level VLM reasoning from low-level motion execution.

\subsection{Lightweight Model Adaptation}
\subsubsection{Visual Semantic Embeddings}
Goal2Pixel contains navigation-specific visual patterns beyond ordinary RGB content, including auxiliary directive regions in the current observation and trajectory overlays in the history memory. Since these regions encode explicit navigation semantics but may not be reliably distinguished by a pretrained vision encoder, we introduce lightweight learnable visual semantic embeddings for the corresponding special visual tokens, shown in the bottom-left part of Fig. \ref{fig:pipeline}. Specifically, we add a learnable directive embedding to current-image tokens that overlap with auxiliary directive regions, and a learnable trajectory embedding to history-image tokens that contain trajectory overlays. All regular RGB tokens and non-trajectory history tokens remain unchanged.

\subsubsection{Training Objective}

Goal2Pixel predicts the navigation target as a coordinate string in the format ``XXX, YYY'', and is primarily trained with the standard token-level cross-entropy loss $\mathcal{L}_{\text{CE}}$. 
However, token-level supervision treats each digit as an independent categorical label, and therefore does not explicitly capture the geometric and metric structure of pixel coordinates. 
To better align supervision with pixel navigation, we introduce two coordinate-aware auxiliary losses.

\paragraph{Soft Predicted Pixel Coordinate}
Since each decimal digit is represented as an individual token by the tokenizer, the ``XXX, YYY'' output contains six explicit digit prediction positions. During teacher-forced training, we derive a differentiable soft coordinate from the per-digit generation logits as an auxiliary training signal. Let $z_k \in \mathbb{R}^{10}$ denote the logits at the $k$-th digit position over digit tokens $\{0,\ldots,9\}$.
We compute the digit distribution and its expected value as
\begin{equation}
\pi_k(d)
=
\frac{\exp(z_{k,d})}
{\sum_{j=0}^{9}\exp(z_{k,j})},
\qquad
\hat{d}_k
=
\sum_{d=0}^{9} d\,\pi_k(d).
\end{equation}
For the coordinate format ``$\hat{d}_1\hat{d}_2\hat{d}_3, \hat{d}_4\hat{d}_5\hat{d}_6$'', the soft predicted coordinate is then composed as
\begin{equation}
\hat{u}
=
100\hat{d}_1
+
10\hat{d}_2
+
\hat{d}_3,
\qquad
\hat{v}
=
100\hat{d}_4
+
10\hat{d}_5
+
\hat{d}_6.
\end{equation}

Before computing the auxiliary losses, we normalize the pixel coordinates by the maximum coordinate value.
Since both $u$ and $v$ are defined in the range $[0,999]$, we compute $(u_n,v_n)=(u/999, v/999)$ and $(\hat{u}_n,\hat{v}_n)=(\hat{u}/999, \hat{v}/999)$, mapping both predicted and ground-truth coordinates to $[0,1]$. During waypoint back-projection, coordinates in $[0,999]$ are linearly mapped back to the $272\times272$ image space.

\paragraph{Numeric and Angular Losses}
Given the predicted pixel $(\hat{u}_n,\hat{v}_n)$ and the GT pixel $(u_n,v_n)$, we define
\begin{equation}
\begin{aligned}
\mathcal{L}_{\text{num}}
&=
\left\|
(\hat{u}_n,\hat{v}_n)
-
(u_n,v_n)
\right\|_1,\\
\mathcal{L}_{\text{ang}}
&=
1-\cos\left(
\theta(\hat{u}_n,\hat{v}_n)
-
\theta(u_n,v_n)
\right).
\end{aligned}
\end{equation}
where $\theta(u_n,v_n)=\operatorname{atan2}(v_0-v_n, u_n-u_0)$ is the egocentric direction from the bottom-center anchor $\mathbf{p}_0=(u_0,v_0)=(0.5,1.0)$ to the pixel. The numeric loss is applied to all coordinate targets, while the angular loss is applied to navigable-pixel and turning targets but masked out for \texttt{Stop}, for which directional error is not meaningful.

\paragraph{Final Training Objective}
The final training objective is defined as
\begin{equation}
\mathcal{L}
=
\mathcal{L}_{\text{CE}}
+
\lambda_{\text{num}}\mathcal{L}_{\text{num}}
+
\lambda_{\text{ang}}\mathcal{L}_{\text{ang}},
\end{equation}
where $\lambda_{\text{num}}$ and $\lambda_{\text{ang}}$ control the contributions of the numeric and angular losses, respectively.

\section{Experimental Results}

\subsection{Experiment Setup}

For \textbf{training}, Goal2Pixel is initialized from InternVL3~\cite{internvl} and fine-tuned for one epoch on our 2.64M-sample VLN-CE corpus from R2R-CE~\cite{r2r-ce} and RxR-CE~\cite{rxr}. We fully fine-tune the LLM backbone and projection layers while freezing the vision encoder. The learning rate is $2\times10^{-5}$, with at most 8 history images. The numeric and angular loss weights are 0.3 and 0.03. The 2B and 8B models require 70 and 176 H100 GPU hours, respectively. 
For \textbf{evaluation}, we report standard VLN-CE metrics on unseen validation splits: Navigation Error (NE), Oracle Success Rate (OS), Success Rate (SR), Success-weighted Path Length (SPL), and normalized Dynamic Time Warping (nDTW). We report results on the full RxR-CE Val-Unseen split for the main comparison, while all RxR-CE ablation studies are conducted on the en-US subset.
For \textbf{real-world evaluation}, we deploy Goal2Pixel on a Mecanum wheeled robot~\cite{mecanum}, running onboard on a laptop with an NVIDIA RTX-5060 GPU. LiDAR measurements are projected into the camera frame to construct the depth image, with interpolation applied to pixels lacking valid depth measurements. We use FAR Planner~\cite{farplanner} as the local planner to track the predicted 3D waypoints and execute robot motion.

\subsection{Main Results}

\begin{table*}[h]
\caption{Comparison with SOTA methods on R2R-CE and RxR-CE Val-Unseen. For VLM-based methods, we also report backbone size. External data refers to training data beyond the official R2R-CE and RxR-CE splits, e.g., EnvDrop~\cite{envdrop}, DAgger~\cite{dagger}, and general VQA data. Results are taken from the respective papers. \colorbox{best}{\textbf{Bold}} and \colorbox{second}{\underline{underline}} denote the best and second-best results. } 
\centering
\small
\setlength{\tabcolsep}{4pt}
\renewcommand{\arraystretch}{1.1}
\resizebox{0.85\textwidth}{!}{%
\begin{tabular}{l|c|cccc|cccc|c}
\toprule
\multirow{2}{*}{Method}
& \multirow{2}{*}{\shortstack{Model\\Size}}
& \multicolumn{4}{c|}{R2R-CE Val-Unseen}
& \multicolumn{4}{c|}{RxR Val-Unseen}
& \multirow{2}{*}{\shortstack{Training\\External Data}} \\
& 
& NE$\downarrow$ & OS$\uparrow$ & SR$\uparrow$ & SPL$\uparrow$
& NE$\downarrow$ & SR$\uparrow$ & SPL$\uparrow$ & nDTW$\uparrow$
& \\
\midrule

InstructNav {\conf{CoRL24}} \cite{instructnav}
& - & 6.89 & -    & 31.0 & 24.0 & -    & -    & -    & -    & - \\
AO-Planner {\conf{AAAI25}} \cite{AO-Planner}
& - & 5.55 & 59.0 & 47.0 & 33.0 & 7.06 & 43.3 & 30.5 & 50.1 & - \\
LaViRA {\conf{ICRA2026}} \cite{LaViRA}
& - & 6.54 & 48.7 & 38.3 & 28.3 & -    & -    & -    & -    & - \\
CA-Nav {\conf{TPAMI2025}} \cite{CA-Nav}
& - & 7.58 & 48.0 & 25.3 & 10.8 & -    & -    & -    & -    & - \\

\midrule
CMA {\conf{CVPR22}} \cite{CMA}
& - & 6.20 & 52.0 & 41.0 & 36.0 & 8.76 & 26.5 & 22.1 & 47.0 & - \\
VLN$\circlearrowright$BERT {\conf{CVPR22}} \cite{CMA}  & - 
& 5.74    & 53.0   & 44.0    & 39.0      & 8.98 & 27.0 & 22.6 & 46.7 & - \\
Ego$^2$-Map {\conf{ICCV23}} \cite{EgoMap}
& - & 5.54 & 56.0 & 47.0 & 41.0 & -    & -    & -    & -    & - \\
g3D-LF {\conf{CVPR25}} \cite{g3d-lf}
& - & 5.70 &  \cellcolor{second}\underline{59.5} & 47.2 & 34.6 & -    & -    & -    & -    & - \\
Sim2Real {\conf{CoRL24}} \cite{sim2real}
& - & 5.95 & 55.8 & 44.9 & 30.4 & 8.79 & 36.7 & 25.5 & 18.1 & - \\
NavMorph {\conf{ICCV25}} \cite{Navmorph}
& - & 5.75 & 56.9 & 47.9 & 33.2 & 8.85 & 30.8 & 22.8 & 44.2 & - \\

\midrule
MapNav {\conf{ACL25}} \cite{Mapnav}
& 7B & 4.93 & 53.0 & 39.7 & 37.2 & 7.62 & 32.6 & 27.7 & 43.5 & 0K \\
NaVid {\conf{RSS24}} \cite{navid}
& 7B & 5.47 & 49.1 & 37.4 & 35.9 & 8.41 & 23.8 & 21.2 & -    & 953K \\
NaVid-4D {\conf{ICRA25}} \cite{NaVid-4D}
& 7B & 5.99 & 55.7 & 43.8 & 37.1 & -    & -    & -    & -    & - \\
StreamVLN {\conf{arXiv25}} \cite{Streamvln}
& 7B & 6.05 & 53.8 & 45.5 & 41.6 & 6.72 & 48.6 & 42.5 & 60.2 & 0K/10033K \\
Uni-NaVid {\conf{RSS25}} \cite{Uni-navid}
& 7B & 5.58 & 53.3 & 47.0 & 42.7 & \cellcolor{best}\textbf{6.24} & 48.7 & 40.9 & -    & 3577K \\
NaVILA {\conf{RSS25}} \cite{Navila}
& 7B & 5.37 & 57.6 & 49.7 & 45.5 & 6.77 & 49.3 & 44.0 & 58.8 & 13132K \\
Efficient-VLN {\conf{arXiv26}} \cite{efficientVLN}
& 7B & 6.41 & 54.5 & 45.9 & 41.9 & 6.51 & \cellcolor{second}\underline{49.8} & 41.5 & 59.4 & 0K \\
JanusVLN {\conf{ICLR26}} \cite{Janusvln}
& 7B & 5.17 & 58.0 & 52.8 & 49.2 & \cellcolor{second}\underline{6.46} & \cellcolor{best}\textbf{51.4} & \cellcolor{second}\underline{44.3} & 59.1 & 0K \\
\midrule
\rowcolor{gray!15}
Goal2Pixel (ours)
& \cellcolor{best}\textbf{2B}
& \cellcolor{second}\underline{4.85} & \cellcolor{best}\textbf{59.9} & \cellcolor{best}\textbf{54.1} & \cellcolor{second}\underline{52.5}
& 7.46 & 42.7 & 39.5 & \cellcolor{second}\underline{60.7}
& 0K \\

\rowcolor{gray!15}
Goal2Pixel (ours)
& 8B 
& \cellcolor{best}\textbf{4.80} & 58.3 & \cellcolor{second}\underline{53.9} & \cellcolor{best}\textbf{52.7}
& 6.91 & 48.1 & \cellcolor{best}\textbf{44.7} & \cellcolor{best}\textbf{63.0}
& 0K \\

\bottomrule
\end{tabular}
}
\label{tab:r2r_rxr}
\end{table*}

\begin{figure*}[h]
    \centering
    \includegraphics[width=0.95\linewidth]{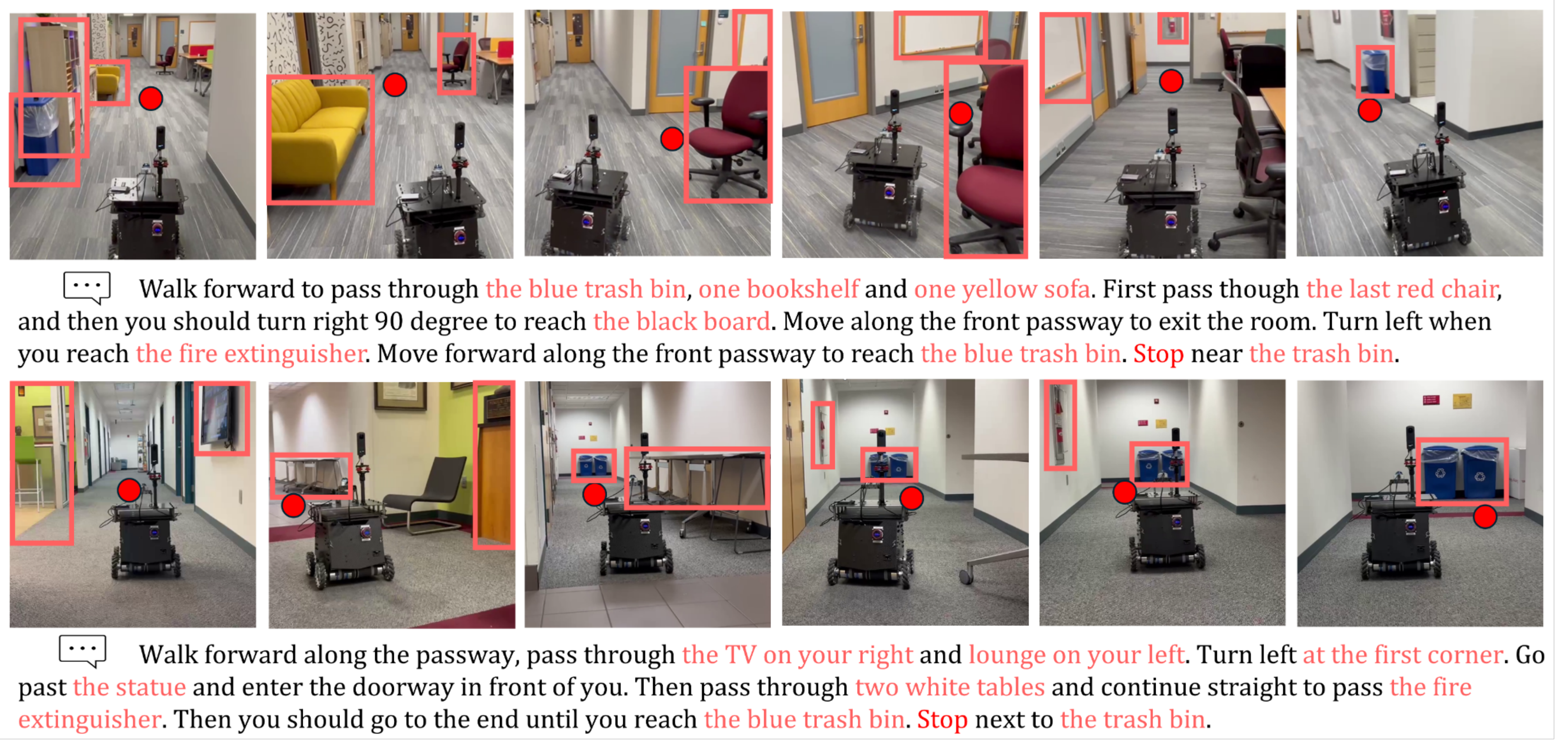}
    \caption{Real-world Goal2Pixel navigation following language instructions. We conduct 30 trials and achieve 16 successful episodes (53.3\% SR). Each image corresponds to one VLM decision step. Pink boxes highlight instruction-relevant landmarks, and red dots indicate the predicted target pixels.}
    \label{fig:real_world_demo}
    \vspace{-5mm}
\end{figure*}

\begin{table*}[h]
\caption{
Ablation study of output paradigms on R2R-CE and RxR-CE.
All output paradigms are defined in Section \ref{sec:output_paradigm}.
\# Calls denotes the average number of VLM invocations per episode, and Inf.~T denotes the average inference time per call.
}
\vspace{3pt}
\centering
\footnotesize
\setlength{\tabcolsep}{3pt}
\renewcommand{\arraystretch}{1.1}
\resizebox{\linewidth}{!}{%
\begin{tabular}{l|l|cccc|cccc}
\toprule
\multirow{2}{*}{\textbf{Output Paradigm}} 
& \multirow{2}{*}{\textbf{Example Output}} 
& \multicolumn{4}{c|}{\textbf{R2R-CE Val-Unseen}} 
& \multicolumn{4}{c}{\textbf{RxR-CE Val-Unseen (US subset)}} \\
& 
& \textbf{SR $\uparrow$} 
& \textbf{SPL $\uparrow$} 
& \textbf{\# Calls $\downarrow$}
& \textbf{Inf. T $\downarrow$}
& \textbf{SR $\uparrow$} 
& \textbf{SPL $\uparrow$} 
& \textbf{\# Calls $\downarrow$}
& \textbf{Inf. T $\downarrow$} \\
\midrule
One Action           
& {\scriptsize One of \texttt{\{Forward/Left/Right/Stop\}}}         
& 32.9 & 31.5 & 46.62 & \cellcolor{best}\textbf{0.067}
& 23.3 & 21.4 & 201.3 & \cellcolor{best}\textbf{0.075}\\

Four Actions       
& {\scriptsize Four from \texttt{\{Forward/Left/Right/Stop\}}}    
& 37.0 & 36.0 & 15.77 & 0.149
& 28.3 & 25.2 & 24.05 & 0.157 \\

\midrule
Action-Pixel (Par)    
& {\tiny\texttt{Forward,XXX,YYY/Left/Right/Stop}}       
& 2.8 & 2.1 & \cellcolor{best}\textbf{4.82} & 0.175
& 8.5 & 7.82 & \cellcolor{best}\textbf{4.03} & 0.179 \\

Action-Pixel (Seq)
& \makecell[l]{%
{\tiny\texttt{Round 1: Select\_Pixel/Left/Right/Stop}}\\
{\tiny\texttt{Round 2: XXX,YYY if Select\_Pixel}}}
& \cellcolor{second}\underline{43.7}
& \cellcolor{second}\underline{41.9}
& 16.85
& 0.161
& \cellcolor{second}\underline{30.1}
& \cellcolor{second}\underline{28.3}
& 17.76
& 0.158\\

\midrule
Pixel  
& {\scriptsize\texttt{XXX, YYY}}                        
& \cellcolor{best}\textbf{54.1} 
& \cellcolor{best}\textbf{52.5} 
& \cellcolor{second}\underline{7.75} 
& \cellcolor{second}0.120
& \cellcolor{best}\textbf{43.8} 
& \cellcolor{best}\textbf{40.4}
& \cellcolor{second}\underline{11.76}
& \cellcolor{second}\underline{0.134}\\
\bottomrule
\end{tabular}%
}
\label{tab:ablation_output}
\end{table*}

\subsubsection{Results on Benchmark R2R-CE and RxR-CE}
Table~\ref{tab:r2r_rxr} shows that Goal2Pixel achieves competitive state-of-the-art performance on both R2R-CE and RxR-CE, compared with prior methods that do not use external training data.
Notably, Goal2Pixel demonstrates strong trajectory quality and path efficiency across both benchmarks.
On R2R-CE, the 8B model achieves 52.7\% SPL, improving +3.5\% over JanusVLN~\cite{Janusvln}. Even with a smaller 2B backbone, Goal2Pixel outperforms prior methods, reaching 54.1\% SR and 52.5\% SPL under the setting not using any external training data.
On RxR-CE, the 8B model achieves 44.7\% SPL and 63.0 nDTW, improving over the strongest prior results by 0.4 and 2.8 points, respectively.
These improvements in SPL and nDTW indicate that Goal2Pixel follows more efficient and  better-aligned trajectories.
We attribute these gains in path efficiency and trajectory alignment partly to the pixel-based interface. Pixel allows the agent to execute more direct local motion toward visible navigable targets, reducing unnecessary oscillations and improving trajectory efficiency.

\subsubsection{Pixel Grounding Quality} 
We further directly evaluate the geometric quality of Goal2Pixel predictions on 104K Val-Unseen samples. The predicted pixels achieve an average pixel L2 error of 37.8 px, corresponding to a 1.20 m 3D waypoint error and a 9.03$^\circ$ angular error after back-projection, while only 7.49\% of predictions fall on non-navigable regions. The relatively small angular error indicates that the predicted pixels generally provide accurate local navigation directions, even when they do not exactly recover the oracle waypoint.

\subsubsection{Real-World Evaluation} We zero-shot deploy Goal2Pixel on a wheeled robot and conduct 30 real-world navigation trials, with 16/30 successful episodes (53.3\% SR). Fig.~\ref{fig:real_world_demo} shows two representative successful trajectories. Without any real-world fine-tuning, Goal2Pixel grounds language instructions into target pixels in real observations and executes the resulting 3D waypoints on a physical robot. Most failures arise from sim-to-real perception gaps and inaccurate depth obtained from LiDAR projection and interpolation, which can lead to incorrect waypoint back-projection. These results demonstrate the feasibility of zero-shot pixel grounding for real-world robot navigation.

\subsection{Ablation Study}

\subsubsection{Output Paradigm} \label{sec:output_paradigm}
Table~\ref{tab:ablation_output} compares five output paradigms with the same backbone, training data, and history representation. One Action and Four Actions predict one or four low-level actions per call, respectively. Action-Pixel (Par) outputs either Forward, XXX, YYY or a discrete non-forward action. Action-Pixel (Seq), following DualVLN~\cite{dualvln}, first predicts pixel selection or a discrete action and invokes a second VLM call for pixel prediction when selected. In contrast, Pixel represents all decisions with a single coordinate pair \texttt{XXX, YYY}, yielding Goal2Pixel's pure Pixel paradigm.

Compared with the DualVLN-style Action-Pixel (Seq) paradigm, Goal2Pixel improves SR/SPL from 43.7/41.9 to 54.1/52.5 on R2R-CE and from 30.1/28.3 to 43.8/40.4 on RxR-CE, while reducing VLM calls from 16.85 to 7.75 and from 17.76 to 11.76, respectively. Goal2Pixel also substantially outperforms Action-Pixel (Par) on both benchmarks. Its poor performance may result from heterogeneous output lengths: forward decisions generate \texttt{Forward, XXX, YYY}, whereas turning and stopping require only one token, creating imbalanced supervision and unmatched semantics. 
Overall, a unified pixel-coordinate representation of heterogeneous navigation decisions provides a more consistent output formulation, avoiding both mixed-format autoregressive supervision and sequential switching between discrete action prediction and pixel grounding.

Compared with direct low-level action prediction, the complete Goal2Pixel system further improves SR by 21.2 percentage points on R2R-CE while reducing the average number of VLM calls from 46.62 to 7.75.
This larger system-level gain reflects both waypoint-level spatial prediction and temporally abstracted execution.

\subsubsection{Execution Horizon}
We further study the effect of the number of low-level execution steps allowed after each VLM prediction.
As shown in Table~\ref{tab:ablation_execution_steps}, increasing the execution horizon from one to five steps substantially improves navigation performance while reducing the number of VLM calls.
On R2R-CE, SR increases from 36.1 at $t=1$ to 54.1 at $t=5$, while the average number of calls decreases from 31.9 to 7.75.
A similar trend is observed on RxR-CE.

\begin{table}[h]
\centering
\footnotesize
\caption{Ablation study of the maximum number of low-level execution steps per VLM prediction on R2R-CE and RxR-CE Val-Unseen (US subset). Avg. \# Calls denotes the average number of VLM invocations per episode.}
\label{tab:ablation_execution_steps}
\setlength{\tabcolsep}{3.5pt}
\renewcommand{\arraystretch}{1.1}
\resizebox{0.9\linewidth}{!}{%
\begin{tabular}{l|ccc|ccc}
\toprule
\multirow{2}{*}{\textbf{Max Exec. Steps $t$}} 
& \multicolumn{3}{c|}{\textbf{R2R-CE Val-Unseen}} 
& \multicolumn{3}{c}{\textbf{RxR-CE Val-Unseen}} \\
& \textbf{SR $\uparrow$} 
& \textbf{SPL $\uparrow$}
& \textbf{Calls $\downarrow$}
& \textbf{SR $\uparrow$} 
& \textbf{SPL $\uparrow$} 
& \textbf{Calls $\downarrow$} \\
\midrule

$t=1$
& 36.08 & 35.66 & 31.90
& 25.02 & 24.22 & 33.48 \\

$t=2$
& 42.37 & 41.65 & 14.82
& 30.09 & 28.85 & 16.92 \\

$t=5$
& \cellcolor{best}\textbf{54.09}
& \cellcolor{best}\textbf{52.45}
& 7.75 
& \cellcolor{best}\textbf{43.83}
& \cellcolor{best}\textbf{40.36}
& 11.76 \\

$t=10$
& \cellcolor{second}\underline{53.57}
& \cellcolor{second}\underline{51.74}
& \cellcolor{second}\underline{4.26}
& \cellcolor{second}\underline{42.44}
& \cellcolor{second}\underline{39.09}
& \cellcolor{second}\underline{6.59} \\

Unlimited
& 40.54 
& 38.46 
& \cellcolor{best}\textbf{2.54}
& 34.26 
& 30.75 
& \cellcolor{best}\textbf{3.87} \\

\bottomrule
\end{tabular}%
}
\end{table}

However, longer execution is not monotonically beneficial.
Increasing the horizon from five to ten steps provides no further performance gain, while unlimited execution substantially reduces SR to 40.5 on R2R-CE and 34.3 on RxR-CE.
This suggests that Goal2Pixel benefits from an intermediate execution horizon. Overly frequent predictions lead to short-horizon decisions and higher VLM cost, whereas overly long execution delays visual feedback and goal correction.

\subsubsection{History Image Representation}
Table~\ref{tab:ablation_history} compares ViKeyMem with commonly used history construction strategies, with all methods limited to at most 8 history images per input. Compared with fixed-interval and different-frequency sampling, ViKeyMem consistently achieves higher SR/SPL on both R2R-CE and RxR-CE while requiring substantially less training time. It significantly outperforms the setting without history images, confirming the importance of historical visual context.
These results suggest that simple temporal sampling may retain redundant observations or miss navigation-relevant intermediate views, whereas visibility-aware keyframe selection preserves more informative navigation history.

Compared with the stronger Uniform History baseline, ViKeyMem achieves a better performance--efficiency trade-off. It reduces training cost from 156 to 70 H100 GPU hours (55\%) and per-call inference time from 0.224s to 0.121s (46\%). Although Uniform History performs slightly better on R2R-CE (56.1/55.1 vs.~54.1/52.5 SR/SPL), ViKeyMem performs better on the longer-horizon RxR benchmark, improving SR/SPL from 41.6/38.5 to 43.8/40.4. This suggests that selectively preserving navigation-relevant frames becomes more beneficial as trajectories grow longer. We further evaluate the trajectory overlay in ViKeyMem. Removing it decreases SR from 54.1 to 48.7 on R2R-CE and from 43.8 to 41.1 on RxR-CE with nearly unchanged computation, indicating that explicitly visualizing the traversed trajectory helps connect sparse keyframes and provides useful motion context to the VLM.

\begin{table}[h]
\centering
\footnotesize
\caption{
Ablation study on history image representations.
Inf. T. and Train T. denote the average inference time (seconds) per call
and H100 GPU training hours, respectively.
}
\label{tab:ablation_history}

\setlength{\tabcolsep}{3pt}
\renewcommand{\arraystretch}{1.1}

\resizebox{\columnwidth}{!}{%
\begin{tabular}{l|cc|cc|cc}
\toprule
\multirow{2}{*}{\textbf{Model Configuration}} 
& \multicolumn{2}{c|}{\textbf{R2R-CE}} 
& \multicolumn{2}{c|}{\textbf{RxR (US subset)}} 
& \multicolumn{2}{c}{\textbf{Time}} \\

& \textbf{SR $\uparrow$} 
& \textbf{SPL $\uparrow$}
& \textbf{SR $\uparrow$} 
& \textbf{SPL $\uparrow$}
& \textbf{Inf. T. $\downarrow$}
& \textbf{Train T. $\downarrow$} \\
\midrule

w/o History Images 
& 34.7 & 33.2 
& 30.6 & 28.3 
& \cellcolor{best}\textbf{0.116} 
& \cellcolor{best}\textbf{36} \\

\midrule

Fixed-Interval (1-step) 
& 37.2 & 35.5 
& 34.1 & 30.2 
& 0.210 & 183 \\

Fixed-Interval (5-step)  
& 46.1 & 45.1 
& 38.7 & 35.4 
& 0.243 & 173 \\

Uniform History 
& \cellcolor{best}\textbf{56.1} 
& \cellcolor{best}\textbf{55.1}  
& \cellcolor{second}\underline{41.6} 
& \cellcolor{second}\underline{38.5} 
& 0.224 & 156 \\

Different Frequency 
& 25.0 & 19.0  
& 24.9 & 21.1 
& 0.242 & 146 \\

\midrule

\textbf{ViKeyMem} 
& \cellcolor{second}\underline{54.1} 
& \cellcolor{second}\underline{52.5} 
& \cellcolor{best}\textbf{43.8} 
& \cellcolor{best}\textbf{40.4} 
& \cellcolor{second}\underline{0.121} 
& \cellcolor{second}\underline{70} \\

\hspace{0.15cm}w/o Trajectory Overlay 
& 48.7 & 47.0 
& 41.1 & 38.1 
& 0.123 & 72 \\

\bottomrule
\end{tabular}%
}
\end{table}

\subsubsection{Lightweight Adaptations} \label{sec:lightweight_adaptation}
Table~\ref{tab:ablation_component} shows that each lightweight adaptation contributes to pixel-grounded navigation. 
Removing the angular or numeric loss consistently reduces SR on both benchmarks, from 54.1 to 53.4/52.3 on R2R-CE and from 43.8 to 42.3/42.9 on RxR-CE. 
This suggests that coordinate-aware auxiliary losses strengthen pixel grounding by helping the model capture the geometric structure of coordinate outputs. 
Removing the visual semantic embeddings also leads to slight performance degradation in most cases, indicating that distinguishing directive regions and trajectory overlays from regular RGB tokens facilitates adaptation to navigable pixel grounding.

\begin{table}[h]
\centering
\footnotesize
\caption{
Ablation study on coordinate-aware losses and visual semantic embeddings.
}
\label{tab:ablation_component}

\setlength{\tabcolsep}{3pt}
\renewcommand{\arraystretch}{1.1}

\resizebox{0.85\columnwidth}{!}{%
\begin{tabular}{l|cc|cc}
\toprule

\multirow{2}{*}{\textbf{Model Configuration}} 
& \multicolumn{2}{c|}{\textbf{R2R-CE}} 
& \multicolumn{2}{c}{\textbf{RxR (US subset)}} \\

& \textbf{SR $\uparrow$} 
& \textbf{SPL $\uparrow$}
& \textbf{SR $\uparrow$} 
& \textbf{SPL $\uparrow$} \\

\midrule

\textbf{Full Model (Ours)}             
& \cellcolor{second}\underline{54.1} 
& \cellcolor{best}\textbf{52.5}
& \cellcolor{best}\textbf{43.8}
& \cellcolor{best}\textbf{40.4} \\

\hspace{0.15cm}w/o Angular Loss
& 53.4 
& 51.8 
& 42.3 
& 39.2 \\

\hspace{0.15cm}w/o Numeric Loss
& 52.3 
& 50.5  
& 42.9 
& 39.6 \\

\hspace{0.15cm}w/o Directive Embedding    
& \cellcolor{best}\textbf{54.2}
& \cellcolor{second}\underline{52.4}
& 42.2 
& 39.3 \\

\hspace{0.15cm}w/o Trajectory Embedding 
& 53.2 
& 51.7 
& \cellcolor{second}\underline{43.3}
& \cellcolor{second}\underline{40.0} \\

\bottomrule
\end{tabular}%
}
\end{table}

\section{CONCLUSIONS}

We presented Goal2Pixel, a pure pixel-based paradigm that reformulates VLN-CE into navigable pixel grounding. By predicting a navigable pixel instead, Goal2Pixel provides a spatial interface between high-level VLM reasoning and executable robot motion. Auxiliary directive regions unify all possible action decisions. To support long-horizon navigation and navigable pixel grounding, we introduced ViKeyMem, a visibility-aware keyframe memory, together with two lightweight model adaptations, including semantic embeddings and auxiliary coordinate-aware losses. Experiments on R2R-CE and RxR-CE show that Goal2Pixel achieves competitive performance while substantially reducing VLM inference calls. Overall, our results suggest that the image plane can serve as a simple, unified, and efficient interface for VLM-based vision-and-language navigation.










\bibliographystyle{IEEEtran} 
\bibliography{example}  

\end{document}